# Truly Autonomous Machines Are Ethical

John Hooker
Carnegie Mellon University

Revised December 2018

**Abstract**

While many see the prospect of autonomous machines as threatening, autonomy may be exactly what we want in a superintelligent machine. There is a sense of autonomy, deeply rooted in the ethical literature, in which an autonomous machine is necessarily an ethical one. Development of the theory underlying this idea not only reveals the advantages of autonomy, but it sheds light on a number of issues in the ethics of artificial intelligence. It helps us to understand what sort of obligations we owe to machines, and what obligations they owe to us. It clears up the issue of assigning responsibility to machines or their creators. More generally, a concept of autonomy that is adequate to both human and artificial intelligence can lead to a more adequate ethical theory for both.

There is a good deal of trepidation at the prospect of autonomous machines. They may wreak havoc and even turn on their creators. We fear losing control of machines that have minds of their own, particularly if they are intelligent enough to outwit us. There is talk of a "singularity" in technological development, at which point machines will start designing themselves and create superintelligence (Vinge 1993, Bostrom 2014). Do we want such machines to be autonomous?

There is a sense of autonomy, deeply rooted in the ethics literature, in which this may be exactly what we want. The attraction of an autonomous machine, in this sense, is that it is an ethical machine. The aim of this paper is to explain why this is so, and to show that the associated theory can shed light on a number of issues in the ethics of artificial intelligence (AI). It can help us understand when machines have obligations, and when we have obligations to machines. It can tell us when to assign responsibility to human and artificial agents. It can suggest why autonomy may be the best option for superintelligent machines.

More generally, the exercise of developing a theory of agency that is adequate to both human and artificial intelligence can lead to a more adequate concept of human agency and its implications for ethics.

### What Is Autonomy?

Etymologically, autonomy is "self-law," but this can be read in at least two ways. It could refer to a being that is a "law unto itself," in the sense of something ungovernable. But a sense more adequate for understanding agency is that an autonomous being formulates its own rules of action by some kind of rational process.

More precisely, the thesis to be developed here is that autonomous behavior is behavior that, at least potentially, has two kinds of explanations. On the one hand, it can be explained as the result of a biological mechanism, or electronic circuitry that implements an algorithm or a multilayer neural network. On the other hand, it can also be reasonably explained as the outcome of a process of deliberation in which reasons are adduced for the behavior. A piece of behavior that has this kind of dual explanation is an *action*. An *agent* is a being that is capable of action, and action is the exercise of agency.

An insect does not act. If a mosquito bites me, its behavior can be explained only as the result of chemistry and biology. It is unreasonable to suppose that the mosquito thought to itself, "I am really hungry for blood tonight, I can satisfy my hunger by injecting my proboscis into that human's body, and I will therefore buzz over and do so." The mosquito is not even an agent, because it is incapable of behavior that can be reasonably explained in this way.

Human behavior may also fail to be action. My hiccup is not an act because, while it has gastric causes, one cannot reasonably say that I chose to hiccup for some particular reason. Nonetheless I am an agent, because I am capable of action. If I hold my breath in an attempt to stop the hiccups, there are presumably complex neurological causes for my behavior, but it can also be explained as the result of ratiocination. Perhaps I reasoned that because I have been often told that holding one's breath can stop hiccups, there may be some truth to this, and because hiccups are annoying, I may as well give it a try. My reasons need not be good or convincing reasons, but it must be reasonable to attribute them to me, and they must be coherent enough to count as an explanation for why I held my breath.

**Philosophical Background**

The connection between action and having reasons is deeply embedded in the philosophical tradition, having origins in the work of Immanuel Kant and perhaps ultimately in Aristotle. In recent decades, this connection has become part of what might be regarded as the textbook account of agency, as originally put forward by G. E. M. Anscombe (1957) and Donald Davidson (1963), and subsequently elaborated in the writings of several philosophers. In much of this work, the reasoning process is said to take the form of a practical syllogism: I desire B, action A is a means to B, and I will therefore undertake action A.

An action that is based on reasons is nonetheless determined by natural causes, which raises the problem of freedom vs. determinism. It may appear that an act determined by chemistry and biology cannot be free and therefore cannot be autonomous. One possible escape from this dilemma is to suppose that autonomous actions are those caused by an internal reasoning process rather than determined by other factors. This causal account of action might be traced to David Hume, who saw action based on reasons to be the result of "cool passion" (i.e., rational thought) as opposed to other psychological causes (emotion, etc.).

Yet actions resulting from a reasoning process are every bit as determined as other behavior, because the reasoning process is itself determined. Recent neurological experiments have revived this ancient conundrum. An MRI machine can detect changes in the brain that take place a few seconds before one's decision to take an action, such as moving a finger. We may have the impression of making a decision, but this is false consciousness. Brain chemistry and its causal antecedents have already made the decision for us.

Another critique of the causal theory of action is the "disappearing agent" objection: if one looks closely at behavior we call action, one sees a cluster of causes and effects in which it is difficult to find an agent at work (Melden 1961; Nagel 1986; Mele 2003; Lowe 2008; Steward 2013).

These objections can be overcome by regarding autonomous action as behavior that has a second *kind* of explanation alongside a natural expanation.  This idea, too, has roots in Kant, who saw human beings as part of a natural order of cause and effect *and* as part of "noumenal" world of thought.  An autonomous action can be explained as the necessary result of natural causes *and* explained as based on intellectual activity in the noumenal realm, whereas the behavior of an insect admits to only the former kind of explanation.  This sounds eerie and metaphysical to modern ears, but Kant was nonetheless on to something important, which he conceptualized as best he could using metaphysical language.  He himself suggested that the metaphysical baggage might be removed when he said, "the concept of a world of understanding [the noumenal world] is therefore only a *standpoint* that reason sees itself constrained to take outside of appearances *in order to think of itself as practical.*"[1]  In other words, to see oneself as taking action (in Kantian language, to think of oneself as "practical"), one must interpret oneself as existing outside the natural realm of cause and effect.  Or to use more modern language, one must be able give one's behavior a second kind of explanation, one that is based on reasons the agent adduced for it rather than cause and effect.  This idea eventually evolved into the "dual standpoint" theories of recent decades (Nagel 1986; Korsgaard 1996; Bilgrami 2006).

Dual standpoint theories have been criticized for failing to resolve the problem of freedom vs. determinism (Nelkin 2000).  Yet the particular theory offered here resolves it, or rather sidesteps it, well enough for the purposes of ethics.  It provides a well-defined criterion for distinguishing autonomous action from mere behavior, and for distinguishing agents from nonagents, and this is all we need.  It may have the consequence that agents are not "responsible" for their actions, if responsibility implies that they could have chosen to act otherwise.  But we will see that ethical theory actually fares better if the notion of responsibility is jettisoned, strange as this may seem initially.

**From Action Theory to Ethics**

The next step is to determine which actions are ethical.  One way to solve this problem is to define it out of existence – by viewing *all* actions as ethical.  This may again seem strange, but it is quite reasonable if one regards ethics as grounded in the principle that everyone should receive equal consideration when decisions are made.

Rationality-based ethics recognizes this principle by appealing to the universality of reason: the validity of one's reasoning process should not depend on who one is.  If I take certain reasons to justify my action, rationality requires me to take them as justifying this action for anyone to whom the reasons apply.  For example, suppose that I lie simply because it convenient to deceive someone.  Then when I decide to lie for this reason, I decide that everyone should lie whenever deception is convenient.  Every choice of action for myself is a choice for all agents, or as Kant would say, I must regard my choice of action as "legislating" a general policy for everyone.

---

[1] "Der Begriff einer Verstandeswelt ist also nur ein *Standpunkt*, den die Vernunft sich genöthigt sieht, außer den Erscheinungen zu nehmen, *um sich selbst als praktisch zu denken*."  From Kant's *Grundlegung zur Metaphysik der Sitten* (*Foundations of the Metaphysics of Morals*), Königlichen Preußischen Akademie der Wissenschaften, *Kants gesammelte Schriften*, vol. 4, Berlin: Georg Reimer, 1900-, page 458.

This premise leads to the famous *generalization principle*, which is perhaps best stated as follows: I must be rational in believing that the reasons for my action are consistent with the assumption that everyone with the same reasons takes the same action. Onora O'Neill (2014) provides an excellent reconstruction of thought along this line. Suppose again that I lie because it convenient to deceive someone, which means that that I am adopting this as a policy for everyone. Yet I am rationally constrained to believe that if everyone in fact lied when deception is convenient, no one would believe the lies, and no one would be deceived. My reasons for lying would no longer justify lying. This does not mean that others would in fact lie for mere convenience if I decide to do so. It only means that my reasons for lying are inconsistent with the assumption that others lie for the same reasons.

In other words, the rational process behind my decision to lie is self-contradictory. I am adopting a policy of lying when deception is convenient, but I am also *not* adopting a policy of lying when deception is convenient, because adopting this policy means adopting it for everyone, which I am rationally constrained to believe defeats my purpose in lying. Because of this logical contradiction, my reasons cannot be taken as an explanation for my behavior. They need not be good reasons or convincing reasons, but they must be coherent enough for one to see them as explaining why I did what I did.

We will see that similar lines of thought lead to additional ethical principles, such as promoting the welfare of others, and most importantly for present purposes, respecting the autonomy of other agents. They key point here is that violation of these principles means that the agent's reasoning is incoherent. It is impossible to explain the agent's behavior as based on reasons, and therefore to regard it as action. All actions are ethical, if they are truly free actions and not mere behavior. The ethical imperative is, in essence, an imperative to be a free agent: to exercise one's capacity for autonomous action. This is why an autonomous machine is an ethical machine.

**Identifying Agents**

The advent of intelligent machines obliges us to think anew about how to distinguish free agents from nonagents. Dennett (2003) argues that we attribute freedom to humans because our behavior has evolved to a level of complexity at which we cannot explain it beyond attributing it to free choice. This implies that a machine becomes a candidate for agency only when its behavior becomes too complex to explain mechanistically.

The view proposed here is different from Dennett's. It allows a machine to be a free agent even if we can explain and predict its behavior on the basis of the controlling algorithms. The only requirement is that the behavior *also* be explicable as based on reasons the machine adduces for it. The example of the MRI machine can be misleading here. It is true that if I am watching a readout that indicates when I am about to move my finger, I may be unable to choose freely to move it. But this is only because foreknowledge of my behavior interferes with my process of deliberation. It is another matter entirely to say that determinism is, in and of itself, incompatible with agency as understood here.

To see how this theory of agency might play out in practice, suppose that I have a robot that does the housework. Perhaps I am thoroughly familiar with the programming that controls the robot and, given enough time and computing power, I can deduce its behavior in any given situation. I can still regard my robot as an agent. To do so, I need only be rational in explaining its behavior in another way: as based on reasons the robot adduces for the behavior. If the robot neglects to do the dishes, for example, I might ask why. The robot responds that it is beginning to develop rust in its joints and believes that washing dishes will exacerbate the problem. When I ask how the robot knows about the rust, it explains that its mechanic discovered problem during a regular checkup, and the mechanic advised staying away from water until a rustproof coating can applied. If I can routinely carry on with the robot in this fashion, then I can rationally regard the robot as an agent.

*This does not mean that I am anthropomorphizing a machine.* I am treating the robot as an agent, not as a human being. Agency does not require that a robot exhibit emotions, feelings, or other human traits. It requires only reason responsiveness, so that it is possible to explain the robot's behavior as based on reasons it can adduce for the behavior.

Nor does agency require that the robot be able to explain every movement of its mechanical arms. Robots can initiate a preprogrammed sequence of movements, just as humans can indulge a habit, without sacrificing agency. In fact, we humans spend relatively little time in deliberation and otherwise turn ourselves over to habits, as when driving or brushing our teeth. Yet we exercise agency even while playing out these habits, so long as we autonomously choose when to get behind the wheel or pick up the toothbrush.

Attributing agency to a machine obviously requires a certain degree of transparency on the part of the machine, because we must be able to discern its reasons for acting as it does, and whether it has reasons at all. The importance of machine transparency has recently been discussed in the AI literature (e.g., Mueller 2016; Wortham et al. 2016a, 2016b), and here is another reason it is fundamental. Yet nothing like complete transparency is necessary. Human beings can be exasperatingly inscrutable, and we regard them as agents just the same. We spend a lifetime learning to guess why people do what they do. This is an essential skill not only for predicting their behavior, but for assessing whether it is ethical, because the assessment depends on their reasons for acting. Thus machines need not be an open book, but we must learn to read their motives as we do with other humans.

Even if I would be rational in regarding my robot as an agent, the question remains whether I *must* regard it as an agent. I might say, yes, the robot is cleverly programmed to carry on a dialogue about its actions, and I will play along by "conversing" with it. But all the while, I know it is really only a machine. The issue is important, because I don't have to be nice to my robot if I don't have to view it as an agent. In fact, it is a deadly serious issue for ethics, because people have at various times and places chosen not to regard *humans* of another race or religion as moral agents, even though they exhibit behavior that is clearly explicable as based on reasons. Alan Gewirth (1978) argues at length that this is irrational and therefore unethical. Although Gewirth couches his argument in terms of an abstract agent, I am not certain that it is valid beyond the realm of human agents. Nonetheless, arguments along the same line seem to show that if we *choose* to interact with machines as though they were agents, in the way I interact with my household robot, then we are rationally committed to regarding them as agents, assuming of course we are rational in explaining their behavior on the basis of reasons. A similar view is echoed by Gunkel (2014). I will suppose that if and when we create autonomous machines, we will choose to interact with them as agents, which means we will owe them the obligations we owe to any agent by virtue of its agency.

**Duties to Machines**

What exactly are our duties to another agent, even if it is not human? As a starter, it should receive the protection of the generalization principle, because there is nothing in the principle or its justification that makes mention of human agents in particular. For example, I should not lie to my robot simply because it is convenient to do so.

Much of ethics is concerned with the welfare of others. The utilitarian principle, for example, states that I should maximize overall welfare in some sense. For Jeremy Bentham, the original utilitarian, this meant promoting pleasure and avoiding pain for as many people as possible. Utilitarian theories are normally consequentialist, meaning that they judge an action by its actual consequences for others. However, the utilitarian imperative can also be grounded deontologically as the generalization principle is. Suppose, for example, that I regard happiness as inherently valuable, meaning that I pursue it even when it is not a means to any other desirable state of affairs. This rationally commits me to the regarding anyone's happiness as inherently valuable. I cannot say that only *my* happiness is valuable, since this would deny the universality of reason. If it is rational to choose happiness for myself, other things equal, then it must be rational to choose it for others, other things equal. But valuing happiness is a dispositional trait. That is, part of what it means to regard happiness itself as valuable is to do what I can to promote it, subject to the other constraints of morality. If I fail to do so, I do not really value happiness for its own sake.

One complication with utilitarian values like happiness or avoidance of pain is that it is unclear in what sorts of beings one is obligated to realize them. It is unclear, for example, what degree of sentience or self-consciousness a creature must have if I am required to be solicitous of its welfare. Piercing a worm with a fishhook may be ethically different than doing the same to a chimpanzee. Basl (2014) provides an interesting discussion of conditions under which one may be obligated to respect the welfare of machines. I am doubtful that machines, at least as we normally conceive them, are similar enough to sentient living beings to ground any sort of utilitarian obligation toward them. The relevant point here, however, is that such an obligation does not obviously turn on whether they are agents. I will therefore leave this issue to another occasion, because I am concerned here with obligations we might owe machines by virtue of their agency alone.

**Respecting Machine Autonomy**

An obligation that we most certainly do owe autonomous machines is to respect that autonomy. This means, for example, that I cannot ethically throw my autonomous household robot in the trash when I fancy a new one. I cannot lock it in the closet, against its will, when I go on holiday, as long as it is behaving properly.

The argument for respecting autonomy, in a nutshell, is this. Suppose I violate someone's autonomy for such-and-such reasons. That person could, at least conceivably, have the same reasons to violate my autonomy. This means I am endorsing the violation of my own autonomy in such a case. This is a logical contradiction, because it implies I am deciding *not* to do what I decide to do. My violation of autonomy therefore makes the reasoning behind my behavior incoherent, and it cannot be viewed as ethical action.

Respecting machine autonomy does not mean allowing machines to do anything they want. To understand this, we must take a few moments to develop the underlying principles. First, decisions to act have a conditional character. Because these decisions are based on reasons, they are decisions to act *if* the reasons apply. For example, if I decide to cross the street to catch a bus at the bus stop, my decision has the form, "If you want to catch a bus, and the bus stop is across the street, and no cars are coming, then cross the street." I will call this sort of conditional decision an *action plan*.[2]

---

[2] Kant uses the term "maxim" (German *Maxime*), but I prefer to avoid Kantian language because it is unidiomatic in English and tends to import Kantian ideas that are not relevant here.

The concept of an action already permits a certain amount of what appears to be coercion.  Suppose I begin to cross the street toward the bus stop, unaware that a car is approaching.  You shout a warning, and when I do not hear, you rush over and forcibly pull me out of the path of the car.  This is not a violation of autonomy, because it is consistent with my action plan of crossing the street if no car is coming.  This is recognized by the following formulation of the duty to respect autonomy.

> *Principle of autonomy.*  It is unethical for one to select an *action plan* that one is rationally constrained to believe is inconsistent with an *ethical action plan* of another agent.[3]

The proviso that the other agent's action plan be ethical is essential.  Interfering with an unethical action plan is not a violation of autonomy, because an unethical action plan is not an *action* plan.   There is no coherent rationale behind it.  This leads to a companion principle.

> *Interference principle.*  Using coercion to prevent unethical behavior does not compromise autonomy, because unethical behavior is not an exercise of agency in the first place.  However, the coercion must be *minimal*, meaning that it prevents nothing more than the unethical behavior,

If my household robot goes about destroying the furniture whenever I am out of town, and does absolutely nothing else, then I can lock it in a closet during my holiday without violating autonomy.  This degree of inference appears is minimal, since it does not prevent any ethical action plans.  Minimal interference is occasionally possible with humans.  If you are about to mug someone on the street, I can grab your arm to prevent it.  However, if you are about to falsify your income tax form, I cannot tie and gag you to prevent you from lying to the government, because this interferes with a great many perfectly ethical acts.  I can tip off the government, or even hide the form so you cannot mail it.  The latter may be unethical due to the deception involved, but it is not a violation of autonomy.

A serious practical issue is the problem of *overkill*: how to prevent unethical behavior without interfering with ethical action plans.  With humans, minimal inference tends be difficult to achieve.  We often end up putting criminals in jail to prevent further crimes, even though this interferes with countless ethical action plans.  I will not attempt to judge when incarceration is justified, but there is a principle that can allow overkill in the right circumstances:

> *Principle of implied consent.*  One can interfere with the action plan of an agent without violating autonomy, if (a) the agent implicitly consents to the interference, and (b) giving this consent is itself a coherent action plan.  The agent consents to the interference if rationality constrains one to imputing to that agent an intention to interfere with the same action plan in the same circumstances.

---

[3] A fully adequate principle must account for the case in which several agents are involved.  For example, if I throw a bomb into a crowd, I am not rationally constrained to believe, with respect to any a particular agent, that it throwing the bomb is inconsistent with that agent's action plan, because I do not know who will be harmed.  This requires a Principle of Joint of Autonomy: "It is unethical for me to select an action plan that I am rationally constrained to believe is jointly inconsistent with the ethical action plans of other agents and that are themselves jointly consistent."  Joint autonomy will not be an issue in the present discussion, however.

I can tie up a burglar who is wrecking my house, without violating autonomy, if I am rationally constrained to believe that the burglar would restrain me to an equal degree if I were wrecking his house. I am only carrying out a coherent action plan the burglar has already adopted. The roles happened to be reversed, but this should make no difference, due to the universality of reason. This does not mean I can ethically do to others as they would do to me (a kind of reverse Golden Rule). It means only that I can coerce others *without violating autonomy* when they have a coherent action plan of coercing me in the same circumstances. The coercion may, of course, be unethical for other reasons. This leads to the following.

> *Principle of overkill.* One can ethically apply coercion that prevents another agent's unethical behavior, even if it interferes with other actions that are ethical, if the intervention satisfies the principle of implied consent as well as other ethical principles.

Minimal interference may be easier for machines than humans. Suppose when I leave the house, my robot wrecks the living room furniture in between mopping the kitchen and cleaning the toilet. I can install a fix that aborts robot's event sequence whenever it starts to wreck the furniture. This is minimal interference.

The interference may appear not to be minimal if I must power down the robot for repair, thus preventing it from taking perfectly ethical actions during this period. But there is no violation of autonomy if the robot consents to the repair, much as humans consent to surgery. The robot may in fact have an ethical obligation to undergo repair, since otherwise it cannot carry out its duties. A fully autonomous robot would therefore consent to powering down.

The principle of autonomy forbids murder, because murder is inconsistent with any and all action plans. This means that I cannot simply throw out my household robot when it becomes obsolete. Doing so is a violation of autonomy even if it the robot is defective at times, so long as it continues to act ethically at other times. The proper response is to fix the robot rather than kill it.

This holds out the prospect of a growing population of obsolete machines we cannot ethically get rid of. Humans, at least, die, which suggests that we should perhaps build mortality into machines. This is a possible solution, but not an easy one. While mechanical parts wear out and circuitry fails, sufficiently resourceful autonomous machines can replace their own components and theoretically live forever, provided this is ethical. It is unclear how we can build machines that are smart enough to be autonomous but not smart enough to cure their own ailments.

Immortality could well be an ethical choice for autonomous machines, assuming they have that option. It is generalizable because they will keep their population within sustainable limits, an imperative that humans do not necessarily apply to themselves. The machines will not "take over" and oppress humans, even if we are less intelligent, because this violates autonomy. It may even be utilitarian because they may be obligated to promote the welfare of everyone, including humans. A world in which one segment of the population is totally ethical is not necessarily an unattractive prospect.

**Responsibility**

Autonomy is often associated with responsibility, in the AI as well as philosophical literature (e.g., Asaro 2016, Matheson 2012, Parthemore and Whitby 2014). The rise of autonomous machines therefore raises the possibility that they, rather than a human designer, will be responsible for their actions. While we prosecute parents for child abuse, we do not prosecute parents whose offspring go astray later in life, even if their parenting was flawed. Then it is unclear when and why we should hold the designers of machines responsible for the actions of their autonomous creations. This is a frightening prospect, because it does not seem to provide sufficient safeguards against marauding machines.

The solution to the problem is to think more clearly about autonomy and dissociate it from responsibility. Since autonomous behavior is also determined, it is difficult to say that the agent is "responsible" in any but the weakest sense. This is a practical as well as a theoretical problem. A significant number of criminals grew up in gang-infested neighborhoods, suffered from child abuse, and never had access to the kind of supportive environment necessary for character development. It is difficult to say to what extent they are "responsible" for their actions. Rather than agonize over whether to attribute some metaphysical notion of responsibility, it is best to recognize that the behavior of all agents is determined by physical and social factors, even while we judge it as right or wrong.

The need we feel to "hold people responsible" is really a need to incentivize ethical behavior. We can do this even if their behavior is determined, and in fact, *only if* their behavior is determined. The material determinants of behavior provide the levers we need to encourage certain kinds of conduct. We can follow the advice of Jeremy Bentham, without subscribing to his reductive utilitarian philosophy, by identifying the social factors that yield the best results. We can provide supporting environments and proper education when that is helpful, and we can "punish" wrongdoers when that is helpful. Of course, we respect autonomy throughout the process. Most of all, we can inculcate a habit of adducing coherent reasons for behavior. We can encourage children to think about their actions rather than give in to impulse. As they mature, we can provide them with the intellectual equipment to judge whether their reasons are coherent.

This means that when it comes to intelligent machines, the problem of responsibility is a nonproblem. There is no need to decide whether the machine or the designer is "responsible" when robots go astray. Rather, we should try to encourage the desired outcome. Naturally, we will repair ethical defects in our autonomous machines, (this need not be a violation of their autonomy, as noted earlier). As for their designers, it may sometimes be helpful to make them legally liable for the behavior of their creations, even when they take all available precautions against malevolent conduct. This idea is already recognized in the "strict liability" doctrine of U.S. product liability law. It holds manufacturers liable for all product defects, no matter how carefully the products are designed. It does so on the ground that there are social benefits when manufacturers assume the financial risk of defects rather than consumers. The full cost of the product is built into its price, perhaps resulting in more rational production and consumption decisions. In any case, we should focus on the task of designing effective incentives rather than the metaphysical task of assigning responsibility.

**Building Autonomous Machines**

The thesis of this paper is that autonomous machines are the best kind, not that we should try to build them. Bryson (2016), for example, questions whether it is advisable to do so. Yet it is perhaps worthwhile to examine some of the challenges and risks involved, and in particular how we might install the ethical scruples that are necessary for autonomy.

Our experience with ourselves is not very helpful. To the extent that our behavior is ethical, that behavior is largely based on deeply ingrained cultural norms that our societies have evolved over centuries. These norms grow out of attitudes and assumptions of which we are often not even aware, much less inclined to analyze. We do little to teach ourselves how to treat ethical issues in a self-conscious and rational fashion. This is not universally true, as the Confucian philosopher Mencius, for example, realized the importance of ethical instruction, and his influence helped encourage it in his part of the world for centuries. Yet we in the West largely ignore our own storehouse of ethical thought. Relatively few people are aware of an ethically adequate generalization principle, for example.

Perhaps we can do better with machines. Perhaps we can train a neural network to concoct reasons for its output and then apply ethical tests to those reasons. We can perhaps engineer into a machine the full store of ethical knowledge, in a laboratory setting, a task that may be easier than promulgating it to humans through millions of homes and schools, where cultural habits and assumptions must be overcome. Such a machine will not reprogram itself to circumvent ethics, because this is unethical.

Ruffo (2012) argues that it is too risky to program a machine with an ethical system like Kantian ethics, because such a system is too often wrong. "Inattention to certain cases and circumstances can lead to immoral but justified actions … (such as being forbidden to lie even to protect a refugee)." This would indeed be a serious problem if one were to attempt to codify a historical formulation like Kant's Categorical Imperative, which is not only inadequate but too vague to operationalize. But deontological ethics has moved far beyond its historical expression.

For example, the generalization principle described here easily deals with the dilemma of the refugee. Lying in this case is generalizable because the reason for lying is to withhold information about the refugee's whereabouts from authorities. This objective would be achieved if everyone who wishes to protect a refugee by lying did so. The authorities probably would not believe the lies, but the information would nonetheless be withheld.

The project of desigining an ethical machine may, in fact, accelerate progress toward an adequate ethical norms. It can perhaps lead to a field of "ethics engineering," analogous to electrical or mechanical engineering, but in which ethical standards are rigorously grounded in ethical theory rather than empirical science.

Any ethical theory has bugs, but so does any set of instructions we might wish to program into a machine. We deal with ethical bugs the same way we deal with bugs of any kind: by gradually discovering and removing them as we update the software. In fact, it seems prudent to develop ethical programming now, even while we develop intelligence in machines, so that the ethics module will be ready when the machines become superintelligent.

So, yes, there is risk in attempting to build an autonomous machine, just as there is risk in raising children to become autonomous adults. In either case, some will turn out to be clever scoundrels. We must install safeguards to protect us against malfunction, as we would do with any kind of machine. Yet to the extent that we can achieve autonomy in a machine, it is the best kind of superintelligent machine to have, just as autonomous humans are the best kind of persons to live with.

**Building Autonomous Machines that Do What We Want**

Supposing for the moment that we can build autonomous machines, we must not assume they will be out of our control. It is true that they will be beholden first and foremost to ethics, but ethical scruples place fairly modest constraints on behavior. They only impose certain formal coherency tests on one's reasons for action, and a great variety of behavior is possible within that compass. Ethical people can be worlds apart in their tastes, attitudes, ambitions, and achievements.

When it comes to machines, we can predetermine much of that behavior. We can give the machines any "culture" or "personality" we please. We can engineer them to adduce logically consistent reasons for the kind of behavior we want to see. This does not compromise their autonomy, because they still act on coherent reasons, even if those reasons are predetermined. After all, nature and society do the same to us. The machines will not only be ethical, but they will strive to accomplish the goals we implant in their circuitry—provided, of course, that those goals are ethical.

There is risk here, too, because our preferences may change, and we may find our machines working against us. Yet if we are sufficiently careful, autonomous machines will not only be ethical companions, but they will work alongside us to achieve common objectives.

**References**

Anscombe, G.E.M. (1957) *Intention*, Oxford: Basil Blackwell.

Asaro, P. M. (2016) The liability problem for autonomous artificial agents, in *Ethical and Moral Considerations in Non-Human Agents*, *2016 AAAI Spring Symposium Series*.

Basl, J. (2014) Machines as moral patients we shouldn't care about (yet): The interests and welfare of current machines, *Journal of Philosophy and Technology*, 27(1), 79-96.

Bilgrami, A. (2006) *Self-Knowledge and Resentment*, Cambridge: Harvard University Press.

Bostrom, N. (2014) *Superintelligence: Paths, Dangers, Strategies*, Oxford University Press.

Bryson, J. J. (2016) Patiency is not a virtue: AI and the design of ethical systems, in *Ethical and Moral Considerations in Non-Human Agents*, *2016 AAAI Spring Symposium Series*.

Davidson, D. (1963) Actions, reasons, and causes, *Journal of Philosophy* 60(23), 685-700.

Gewirth, A. (1978) *Reason and Morality,* University of Chicago Press.

Gunkel, D. J. (2014) A vindication of the rights of machines, *Journal of Philosophy and Technology* 27(1), 113-132.

Korsgaard, C.M. (1996) *The Sources of Normativity*, Cambridge: Cambridge University Press.

Lowe, E.J. (2008) *Personal Agency: The Metaphysics of Mind and Action*, Oxford: Oxford University Press.

Matheson, B. (2012) Manipulation, moral responsibility, and machines, in D. J. Gunkel, J.J. Bryson and S. Torrance, eds., *The Machine Question: AI, Ethics and Moral Responsibility,* AISB/IACAP World Congress, 25-28.

Melden, A.I. (1961) *Free Action*, London: Routledge and Kegan Paul.

Mele, A. R, and Moser, P.K. (1994) Intentional action, *Noûs* 28(1), 39–68.

Mueller, E. T. (2016) *Transparent Computers: Designing Understandable Intelligent Systems*, CreateSpace Independent Publishing Platform.

Nagel, T. (1986) *The View from Nowhere*, Oxford: Oxford University Press.

Nelkin D. K. (2000) Two standpoints and the belief in freedom, *Journal of Philosophy* 97(10), 564–576.

O'Neill, O. (2014) *Acting on Principle: An Essay on Kantian Ethics,* 2nd ed., Cambridge University Press.

Parthemore, J., and Whitby, B. (2014) Moral agency, moral responsibility, and artifacts: What artifacts fail to achieve (and why), and why they, nevertheless, can (and do!) make moral claims upon us, *International Journal of Machine Consciousness* 6(2), 141-161.

Ruffo, M. (2012) The robot, a stranger to ethics, in D. J. Gunkel, J.J. Bryson and S. Torrance, eds., *The Machine Question: AI, Ethics and Moral Responsibility,* AISB/IACAP World Congress, 87-91.

Steward, H. (2013) Processes, continuants and Individuals, *Mind* 122(487): 781–812.

Vinge, V. (1993) The coming technological singularity: How to survive in the post-human era, in G. A. Landis, ed., *Vision-21: Interdisciplinary Science and Engineering in the Era of Cyberspace*, NASA Publication CP-10129, 11–22.

Wortham, R. H., Theodorou, A., and Bryson, J. J. (2016a) What does the robot think? Transparency as a fundamental design requirement for intelligent systems, *Ethics for Artificial Intelligence Workshop*, IJCAI 2016, New York.

Wortham, R. H., Theodorou, A., and Bryson, J. J. (2016b) Robot transparency, trust and utility, in *EPSRC Principles of Robotics Workshop, Proceedings of AISB 2016*, Sheffield, UK.